\documentclass{bmvc2k}
\usepackage{times}
\usepackage{epsfig}
\usepackage{graphicx}
\usepackage{amsmath}
\usepackage{amssymb}
\usepackage[ruled]{algorithm2e}
\usepackage{dsfont}
\usepackage{multirow}
\usepackage{booktabs}
\usepackage{wrapfig}

\title{Score-PA: Score-based 3D Part Assembly}

\addauthor{Junfeng Cheng}{junfeng.cheng20@imperial.ac.uk}{1}
\addauthor{Mingdong Wu}{wmingd@pku.edu.cn}{2}
\addauthor{Ruiyuan Zhang}{zhangruiyuan@zju.edu.cn}{3}
\addauthor{Guanqi Zhan}{guanqi@robots.ox.ac.uk}{4}
\addauthor{Chao Wu}{chao.wu@zju.edu.cn}{5}
\addauthor{Hao Dong}{hao.dong@pku.edu.cn}{$\dag$2}

\addinstitution{
 Department of Electrical and Electronic Engineering,\\
 Imperial College London,\\
 London, UK
}
\addinstitution{
 School of CS,\\
 Peking University,\\
 Beijing, China
}
\addinstitution{
 College of Computer Science and Technology,\\
 Zhejiang University,\\
 Hangzhou, China
}
\addinstitution{
 Department of Engineering Science,\\
 University of Oxford,\\
 Oxford, UK
}
\addinstitution{
 School of Public Affairs,\\
 Zhejiang University,\\
 Hangzhou, China
}

\runninghead{Cheng et al.}{Score-PA: Score-based 3D Part Assembly}

\begin{document}

\maketitle

\begin{abstract}

\let\thefootnote\relax\footnotetext{$\dag$ Corresponding Author}

Autonomous 3D part assembly is a challenging task in the areas of robotics and 3D computer vision. This task aims to assemble individual components into a complete shape without relying on predefined instructions. In this paper, we formulate this task from a novel generative perspective, introducing the Score-based 3D Part Assembly framework (Score-PA) for 3D part assembly. Knowing that score-based methods are typically time-consuming during the inference stage. To address this issue, we introduce a novel algorithm called the Fast Predictor-Corrector Sampler (FPC) that accelerates the sampling process within the framework. We employ various metrics to assess assembly quality and diversity, and our evaluation results demonstrate that our algorithm outperforms existing state-of-the-art approaches. We release our code at \url{https://github.com/J-F-Cheng/Score-PA_Score-based-3D-Part-Assembly}.

\end{abstract}

\section{Introduction}
\label{sec:Introduction}

Assuming you purchase a piece of IKEA furniture, assembling the separate parts into a complete structure can be challenging without proper guidance (\emph{e.g.}, the instructions in the manual). If we were to use a robot to assist us with furniture assembly, a key issue would be enabling the robot to understand the relationships among all the parts and autonomously assemble them. Tackling autonomous 3D part assembly is a complex and demanding task in the fields of robotics and 3D computer vision. This challenge arises due to the requirement for algorithms to explore and navigate through a large pose space for each input part, identifying the correct orientation and position for assembly while also accounting for various constraints and dependencies between the components.

Recent years have witnessed significant explorations in the field of 3D part assembly. Dynamic Graph Learning (DGL) \cite{zhan2020generative} is a representative algorithm in this area, featuring an iterative graph neural network designed for the task. This network can predict a 6-DoF part pose for each input part and transform input part point clouds to assemble the shape. RGL-NET \cite{harish2021rgl}, another notable algorithm for 3D part assembly, leverages the order information of input point clouds to enhance assembly capabilities.

According to Huang et al., the 3D part assembly task faces two primary challenges: assembly quality and assembly diversity. The former necessitates that the designed algorithm accurately assemble parts into a complete shape, while the latter requires the algorithm to produce a range of reasonable assembly outcomes. Existing methods, which focus on minimizing the distance between the network-predicted pose and the ground truth pose, struggle to generate diverse results. To address this, we recast the problem as a generative task. Mathematically, we learn a conditional probability $p(\mathbf{Q_P} \mid \mathbf{P})$ to generate new poses, where $\mathbf{Q_P}$ represents the pose set for the input part set $\mathbf{P}$. We propose the Score-based 3D Part Assembly framework (Score-PA), which can learn the proposed conditional distribution. Our experiments demonstrate that our framework attains superior quality and greater diversity in results compared to other baselines. However, the inference stage of a score-based model is typically time-consuming. We thus propose the Fast Predictor-Corrector Sampler (FPC) to accelerate the inference stage of our framework.

Following Huang et al. \cite{zhan2020generative}, we employ Shape Chamfer Distance (SCD) \cite{zhan2020generative}, Part Accuracy (PA) \cite{li2020learning}, and Connectivity Accuracy (CA) \cite{zhan2020generative} to assess the quality of the generated results. However, Huang et al. only provide a qualitative evaluation of algorithm diversity and do not use any quantitative metrics for this aspect. To address this gap, we introduce Quality-Diversity Score (QDS) and Weighted Quality-Diversity Score (WQDS) to compare our algorithm's diversity with that of other baselines. Both qualitative and quantitative results demonstrate that our proposed method significantly surpasses previous algorithms in terms of diversity while achieving state-of-the-art performance for quality.

We summarise our contribution as follows:
\begin{itemize}
    \item We approach the autonomous 3D part assembly task from a novel generative perspective, proposing the Score-based 3D Part Assembly framework (Score-PA). This framework learns a conditional probability for the pose set of input parts, resulting in high-quality and diverse results.
    \item We propose a new sampler called FPC to speed up the convergence of our Score-based 3D Part Assembly framework.
    \item We also propose two new metrics to quantitatively evaluate the assembly diversity. 
\end{itemize}

\vspace{-.5cm}
\section{Related Work}

\subsection{Assembly-based 3D modeling}
Although our work does not focus on 3D modelling, it is necessary to review assembly-based 3D modelling techniques, as they have already proposed assembly methods for creating new 3D models. Assembly-based 3D modelling is a promising approach to expanding the accessibility of 3D modelling. In assembly-based modelling, new models are constructed from shape components extracted from a database \cite{chaudhuri2011probabilistic}. One approach, proposed by Funkhouser et al. \cite{funkhouser2004modeling}, creates new 3D shapes by assembling parts from a repository. Other works \cite{chaudhuri2011probabilistic, kalogerakis2012probabilistic, jaiswal2016assembly} facilitate shape modelling using probabilistic models to encode semantic and geometric relationships among shape components. More recently, some studies \cite{li2020learning, schor2019componet, wu2020pq} generate certain parts and then predict a per-part transformation for the generated parts to obtain a new shape.

\subsection{Autonomous 3D Part Assembly}

The autonomous 3D part assembly task, as introduced by Huang et al. \cite{zhan2020generative}, focuses on predicting a 6-DoF pose (comprising rotation and translation) for the composition of individual parts. To accomplish this, Huang et al. \cite{zhan2020generative} proposed an assembly-oriented dynamic graph learning framework that demonstrated impressive performance using their specially designed algorithm. Following this development, a recurrent graph learning framework was introduced \cite{harish2021rgl}, which takes advantage of the order information of parts during the assembly process. This approach highlights the significant improvements that can be achieved by incorporating order information. However, in practical applications, such order information is often not readily available for 3D part assembly problems. As a result, our study continues to adhere to the setting proposed by Huang et al. \cite{zhan2020generative}, focusing on assembling parts in random order. This ensures that our approach remains applicable in real-world scenarios where order information might not be accessible.

\subsection{Score-Based Generative Models}

Score-based generative models aim to estimate specific distributions \cite{song2020score, song2020improved, hyvarinen2005estimation, song2020sliced, song2021maximum}. The primary objective is to minimize the squared distance between the estimated gradients and the gradients of the log-density of the data distribution \cite{hyvarinen2005estimation, song2019generative}.

A recent variation of score-based models, known as score-based generative modelling with stochastic differential equations (SDEs) \cite{song2020score}, employs SDEs for data perturbation, achieving remarkable success in generation tasks. This approach diffuses the data distribution during training using SDEs and generates data by reversing the diffusion process, \emph{i.e.}, through the inverse SDE.

Score-based modelling has achieved significant success in various tasks, such as point cloud generation \cite{cai2020learning}, molecular conformation generation \cite{shi2021learning}, scene graph generation \cite{suhail2021energy}, point cloud denoising \cite{luo2021score}, human pose estimation \cite{ci2022gfpose}, object rearrangement \cite{wu2022targf}, \emph{etc.}

\section{Method}

\label{section:method}

This section focuses on introducing our proposed approach to tackle the challenges discussed above. We discuss the problem definition in Section \ref{set:problem_formulation}. We then overview our Score-based 3D Part Assembly framework in Section \ref{sec:overview}, and the training algorithm in Section \ref{sec:loss}.
To speed up the inference stage of our framework, we further propose a new sampler, FPC, for fast sampling purpose in Section \ref{sec:noise_decay}.

\subsection{Problem Definition}
\label{set:problem_formulation}

Assuming we have a set of separate parts, the core concept of autonomous 3D part assembly involves designing an algorithm that learns the assembly rules from a dataset and moves the separate parts to appropriate locations accordingly. Specifically, the goal of the 3D part assembly task is to predict a 6-DoF pose set $\mathbf{Q_P}=\{\mathbf{q}_i\}_{i=1}^N$ for parts transformation, corresponding to the given 3D part point clouds $\mathbf{P}=\{\mathbf{p}_i\}_{i=1}^N$ (the order of the input parts is random).
Each $\mathbf{p}_i\in \mathbb{R}^{1000\times 3}$ is a point cloud that conveys the geometric information of a part, and each $\mathbf{q}_i \in \mathbb{R}^6$ is a combination of a translation vector $\mathbf{t}_i \in \mathbb{R}^3$ and an Euler angle vector $\mathbf{e}_i \in \mathbb{R}^3$.
By using the predicted pose set $\mathbf{Q_P}=\{\mathbf{q}_i\}_{i=1}^N$, the given point clouds $\mathbf{P}=\{\mathbf{p}_i\}_{i=1}^N$ can be transformed into an assembled shape $\mathbf{P}^*$ through rotation and translation.

\subsection{Overview of Score-PA}
\label{sec:overview}

\begin{figure}[t]
  \centering
  \includegraphics[width=.99\textwidth,trim=30 150 0 0,clip]{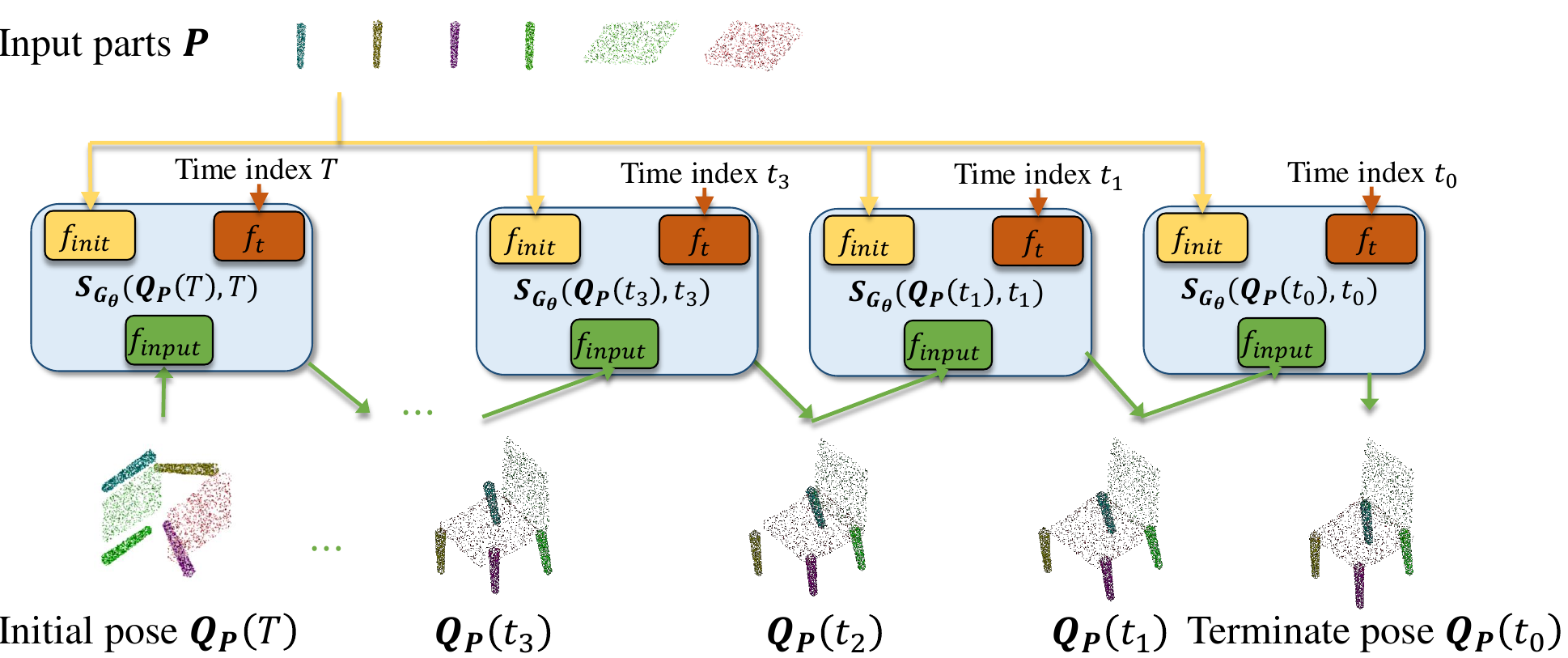}
  \caption{
  The sampling procedure of our Score-PA. We here show a process of $N$ steps sampling. $t_n$ represents the time value at the step $n$ (the algorithm starts from step $N-1$ to step $0$). We can observe that the chair is assembled from coarse to fine. 
  }
  \label{Fig:gen_process}
\end{figure}

As stated in Section \ref{sec:Introduction}, our goal is to learn a conditional probability distribution $p(\mathbf{Q_P} \mid \mathbf{P})$. Instead of directly predicting the final part pose, we learn how to transform the input part by predicting the gradient field that can guide each part to the conditional distribution. Score-based generative model, as an emerging generative technique, provides a direct way to learn the gradient field. Our goal is to build score function to approach the conditional data distribution $\nabla_\mathbf{Q_P} log(p_{t}(\mathbf{Q_P} \mid \mathbf{P}))$. To achieve this, we design a graph neural network to model our score function $\mathbf{S}_{\mathbf{G_\theta}}(\mathbf{Q_P},\ t)=\nabla_\mathbf{Q_P} log(p_{t}(\mathbf{Q_P} \mid \mathbf{P}))$ \cite{song2020score}, where $\mathbf{G_\theta}$ is our designed graph neural network, and time $t$ is used to index the diffusion process of 6-DoF pose $\{\mathbf{Q_P}(t)\}^T_{t=0}$. The overview of our framework is shown in Fig.~\ref{Fig:gen_process}, our objective is to train the formulated score-based model to generate the 6-DoF pose set $\mathbf{Q_P}$ based on the given 3D part point clouds information of the parts $\mathbf{P}$, and $\mathbf{P}$ can be easily transformed to the assembled shape $\mathbf{P}^*$. In the proposed algorithm, the graph $\mathbf{G_\theta}=(\mathbf{V}, \mathbf{E})$ represents the 3D geometric information of the given parts' point clouds. The nodes in this graph neural network are fully connected for message passing. The set of nodes $\mathbf{V}=\{\mathbf{v}_i\}_{i=1}^N$ is obtained via the parts' point clouds $\mathbf{P}=\{\mathbf{p}_i\}_{i=1}^N$ through a parametric function $f_{init}$ which is designed as a vanilla PointNet \cite{qi2017pointnet}. It can extract the geometric information from the parts' point clouds:
\begin{equation}
    \mathbf{V} = f_{init}(\mathbf{P}),
\end{equation}
The \emph{Input Encoder} $f_{input}$ shown in Fig.~\ref{Fig:gen_process} is parameterized as an MLP, which is used to encode the input of the designed network.
The embedding layer $f_t$ includes a \emph{Gaussian Fourier Projection} layer and a linear layer, which enable the score-based model to be conditioned on the input time $t$ \cite{song2020score, tancik2020fourier}.

Subsequently, for training the model, as indicated by Song et al. \cite{song2020score}, training with perturbed data can help the model better estimate the score. In our designed framework, we train our model with the data perturbed by SDEs, and under the certain condition $\mathbf{P}$, a diffusion process of 6-DoF pose $\{\mathbf{Q_P}(t)\}^T_{t=0}$ is constructed for this purpose, where $t \in [0,T]$. $\mathbf{Q_P}(0) \sim p_0$ represents the i.i.d. samples in the given dataset, and $\mathbf{Q_P}(T) \sim p_T$ is a prior distribution that we have already known. The diffusion process can be expressed by the following mathematical model:
\begin{equation}
d \mathbf{Q_P} = \mathbf{f_d}(\mathbf{Q_P}, t) d t + g(t) d \mathbf{w},
\end{equation}
where $\mathbf{f_d}(\cdot, t): \mathbb{R}^d \to \mathbb{R}^d$ represents the drift coefficient of $\mathbf{Q_P}$, and $g(t) \in \mathbb{R}$ is diffusion coefficient \cite{song2020score}. $\mathbf{w}$ is the standard Brownian motion \cite{song2020score}. More details about the training algorithm are discussed in Section \ref{sec:loss}.

After training, we can generate poses for the separate parts through sampling.
Assume a well-trained model is obtained, and we can sample the 6-DoF pose set $\mathbf{Q_P}(0)$ by using reverse-time SDE shown as follows:
\begin{equation}
  d\mathbf{Q_P} = [\mathbf{f_d}(\mathbf{Q_P}, t) - g^2(t)\nabla_{\mathbf{Q_P}}\log (p_t(\mathbf{Q_P} \mid \mathbf{P}))] dt + g(t) d\bar{\mathbf{w}},
\end{equation}
where $\bar{\mathbf{w}}$ represents a reverse time Brownian motion, and $dt$ is an infinitesimal negative time step. In the real scenario, we normally use an iterative algorithm (\emph{e.g.}, Predictor-Corrector sampling algorithm) to solve the inverse SDEs. The simplified sampling process is described in Fig.~\ref{Fig:gen_process}. Assuming we conduct $N$ steps sampling, the iterative algorithm starts with the initial pose $\mathbf{Q_P}(T)$. Then the initialized value is processed by the trained score-based model to obtain $\mathbf{Q_P}(t_{N-2})$. The iterative algorithm continues until the terminate pose $\mathbf{Q_P}(t_0)$ is obtained ($t_0=0$), and the final value is used for translating and rotating the input parts. In our framework, we design a new algorithm, Fast Predictor-Corrector sampler (FPC), to speed up the sampling procedure. More details about our sampling algorithm are discussed in Section \ref{sec:noise_decay}.

\subsection{Loss Function and Training Algorithm}
\label{sec:loss}

Our proposed algorithm aims to estimate the conditional distribution of training data $p(\mathbf{Q_P}(0) \mid \mathbf{P})$, and the original score-matching method is not suitable for this scenario since it is designed for single random variable estimation. We propose a new objective function to solve this problem. Different from the original score-matching objective function \cite{song2020score}, our objective estimates the gradient fields of log-conditional-density $\nabla_{\mathbf{Q_P}} log(p_{t}(\mathbf{Q_P}(t) \mid \mathbf{P}))$. The formula is shown as follows:

\vspace{-.5cm}

\begin{equation}
\begin{aligned}
\label{equ:training_objective}
\min_\theta \mathbb{E}_{t} \mathbb{E}_{\mathbf{Q_P}(0) \mid \mathbf{P}}\mathbb{E}_{\mathbf{Q_P}(t) \mid \mathbf{Q_P}(0), \mathbf{P}}\left[\lambda(t) \Vert\mathbf{S_{G_{\theta}}}(\mathbf{Q_P}(t), t) - \nabla_{\mathbf{Q_P}(t)}\log (p_{0t}(\mathbf{Q_P}(t) \mid \mathbf{Q_P}(0), \mathbf{P}))\Vert_2^2\right]
\end{aligned}
\end{equation}

\begin{wrapfigure}{r}{0.5\textwidth}
\begin{algorithm}[H]

\SetCustomAlgoRuledWidth{0.45\textwidth}  
\caption{\small{Training algorithm of our Score-PA}}
\small
\label{algorithm:training}
\KwIn{Training dataset $\mathcal{D}_{train}$}
\textbf{Parameters}: $T$, $\sigma$

\For{N epochs}{
    \For{each $\mathbf{P}, \mathbf{I_P}, \mathbf{Q_P}(0)$ from the training dataset $\mathcal{D}_{train}$}{
    Sample $t$ from uniform distribution $\mathcal{U}(0, T)$
    
    $\mathbf{Q_P}(t) \leftarrow \mathbf{Q_P}(0) + \mathcal{N}(\mathbf{0}, \frac{1}{2 \log \sigma}(\sigma^{2t} - 1)\mathbf{I})$ 
    Optimize Equation \ref{equ:training_objective}.
    }
}
\end{algorithm}
\end{wrapfigure}

In our algorithm, we set the perturbation SDE as $d \mathbf{Q_P} = \sigma^t d\mathbf{w}$, where $t\in[0,T]$. We select $\lambda(t) = \frac{1}{2 \log \sigma}(\sigma^{2t} - 1)$ in our experiment. Algorithm \ref{algorithm:training} shows our training algorithm. Intuitively, for each iteration, we first select the training data from the training dataset and sample $t$ from the uniform distribution. Then the ground truth pose $\mathbf{Q_P}(0)$ is perturbed by a Gaussian noise with variance $\frac{1}{2 \log \sigma}(\sigma^{2t} - 1)\mathbf{I}$. Finally, we calculate the objective function shown in Equation \ref{equ:training_objective} and use the gradient-descent algorithm to optimize the parameter of the designed graph neural network.

\subsection{Fast Predictor-Corrector Sampler for Inference}
\label{sec:noise_decay}

\begin{wrapfigure}{r}{0.5\textwidth}
\vspace{-2cm}
\begin{algorithm}[H]

\SetCustomAlgoRuledWidth{0.45\textwidth}  

\caption{\small{Fast Predictor-Corrector Sampling Algorithm (FPC)}}
\small
\label{algorithm:FPC}
\KwIn{Testing dataset $\mathcal{D}_{test}$, Graph Neural Network $\mathbf{G_\theta}$}
\textbf{Require}: $T$, $\sigma$, $N$, $d$, $C$, $C_F$, $r$

Select $\mathbf{P}, \mathbf{I_P}$ from the testing dataset $\mathcal{D}_{test}$. 

Sample $\mathbf{Q_P}(T) \sim \mathcal{N}(\mathbf{0}, \frac{1}{2 \log \sigma}(\sigma^{2T} - 1)\mathbf{I})$.

\For{$n \leftarrow N-1$ to $1$}{ \tcp{Original Predictor-Corrector sampling algorithm}
    
	$t_p \leftarrow \frac{(n+1)T}{N}$ $t \leftarrow \frac{nT}{N}$
	
	\For{$i \leftarrow 1$ to $C$}{

	$\mathbf{Q_P}(t_p) \leftarrow Corrector(\mathbf{Q_P}(t_p))$
	}
	$\mathbf{Q_P}(t) \leftarrow Predictor(\mathbf{Q_P}(t_p)) $
}

$t_p \leftarrow \frac{T}{N}$

\For{$i \leftarrow 0$ to $C_F-1$}{
	$\mathbf{z} \sim \mathcal{N}(\mathbf{0}, \mathbf{I})$
	
	$\mathbf{g} \leftarrow \mathbf{S_{G_{\theta}}}(\mathbf{Q_P}(t_p), t_p)$
	
	$\epsilon \leftarrow 2(r ||\mathbf{z}||_2/||\mathbf{g}||_2)$
	
	$\mathbf{Q_P}(t_p) \leftarrow \mathbf{Q_P}(t_p) + \epsilon \mathbf{g} + \sqrt{2\epsilon (1-\frac{i}{C_F})^d} \mathbf{z}$ \tcp{Corrector with noise decay}
	}
	
	$\mathbf{Q_P}(0) \leftarrow \mathbf{Q_P}(t_p) + \frac{T}{N}\sigma^{2t}\mathbf{S_{G_{\theta}}}(\mathbf{Q_P}(t_p), t_p)$ \tcp{Predictor without noise}

\textbf{return} $\mathbf{Q_P}(0)$

\end{algorithm}
\end{wrapfigure}
Assume a well-trained model is obtained by using the training method discussed above, and we can then use this trained model to sample the poses of the separate parts. We first apply the Predictor-Corrector sampler (PC) proposed by Song et al. \cite{song2020score} in our framework to sample poses. However, we find that the PC sampler requires a large number of sampling steps to achieve high performance. Fig.~\ref{fig:ablation_fast_sampling} in Section \ref{sec:ablation_study} shows that the PC sampler requires 400 steps to achieve optimal sampling results. This means it brings a large latency in the inference stage.  Motivated by accelerating the sampling speed, we propose Fast Predictor-Corrector sampler (FPC).
Algorithm \ref{algorithm:FPC} shows the sampling process of our proposed algorithm. Assume we conduct $N$ steps sampling with $C_F$ steps final correction. From step $N-1$ to step $1$, the algorithm executes the same program as the Predictor-Corrector sampler algorithm proposed by \cite{song2020score}. After that, the algorithm conducts $C_F$ steps Langevin MCMC \cite{geyer1992practical, brooks1998markov, brooks2011handbook, carlo2004markov} with noise decay for correction. Finally, a one-step prediction step without noise is used to obtain the results. $d$ is a parameter which controls the decay rate of the noise. The experimental results discussed in Section \ref{sec:ablation_study} show that our algorithm generates better and more diverse results compared with that of the original Predictor-Corrector sampler. 

\paragraph{The design rationale of the FPC}
We initially apply $N-1$ steps of normal PC sampling. The objective of this stage is to find the \textbf{approximate value} of a reasonable pose. At this stage, we refrain from implementing noise decay techniques as they could potentially impede both the \textbf{convergence} of the algorithm and the \textbf{diversity} of the produced results. Then, once the algorithm finds the approximate value of a reasonable pose, we utilize $C_F$ steps of Langevin MCMC to obtain the \textbf{exact value} of the pose. In the part assembly task, even slight noise can corrupt the pose. In this case, we implement the noise decay technique at this stage to avoid corrupting the sampled pose. At this point, the noise decay no longer hinders convergence and diversity (because the algorithm has already found the approximate value), but instead enhances the quality of sampling, which brings a more accurate estimation of the pose. As a result, the algorithm can also achieve better connectivity.

\begin{figure}[t]
  \centering
  \includegraphics[width=1.0\textwidth,trim=40 200 120 10,clip]{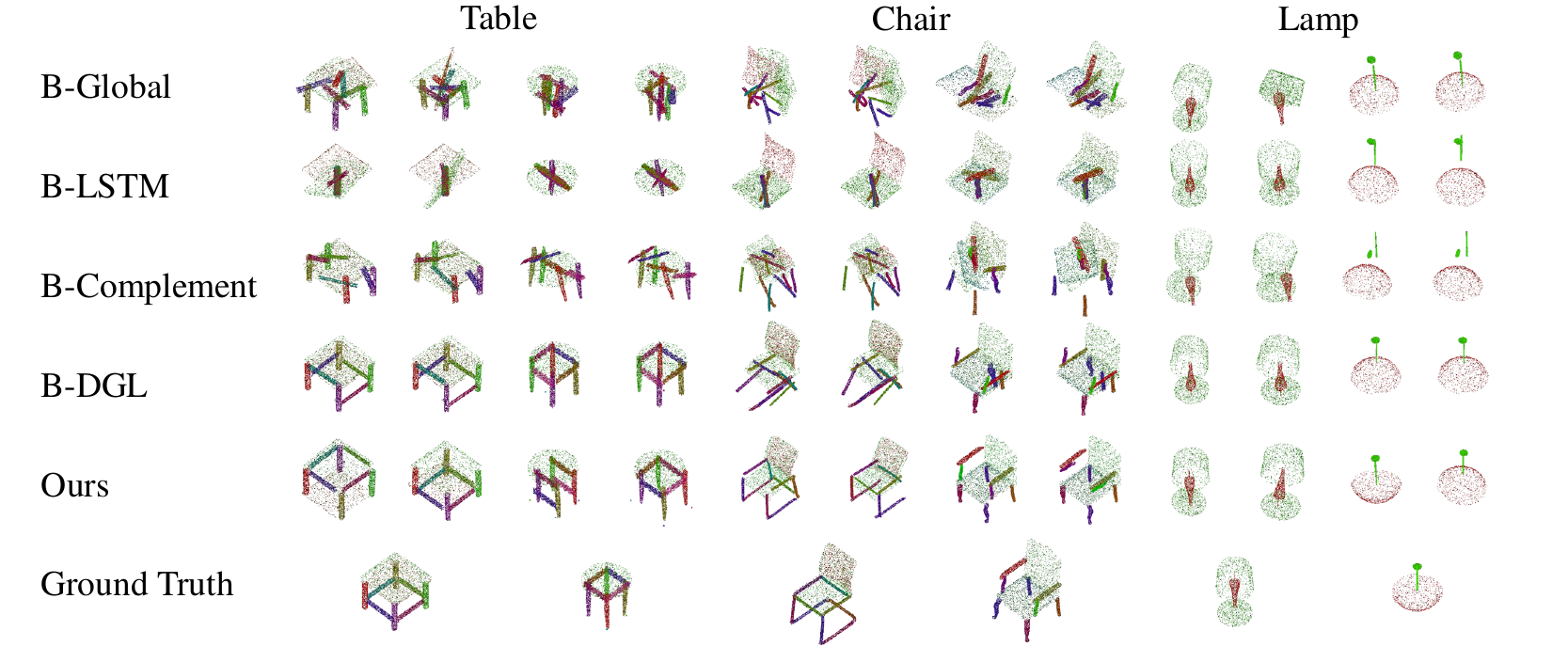}
  \caption{The qualitative comparisons between our algorithm and other baselines. For each algorithm, we show two generated assemblies. The results show that only our framework is able to generate diverse results with high quality. It is hardly possible for B-Global, B-LSTM and B-Complement to generate reasonable results. B-DGL can generate some reasonable assemblies, but the generated assemblies lack diversity. We show more results in the appendix.}
  \vspace{-.4cm}
  \label{fig:eval_qualitative_results}
\end{figure}

\section{Experiment}
\label{sec:Experiment}
In this section, we discuss our datasets, baselines, evaluation metrics, experimental results and ablation study. For the experimental details, please refer to the appendix.
\subsection{Datasets and Baselines}

The datasets used for experiments follow \cite{zhan2020generative}, which consist of three categories: chair, table and lamp (these datasets are subsets of PartNet \cite{mo2019partnet}, which contain a large number of fine-grained shapes and hierarchical part segmentations). Chair dataset, Table dataset and Lamp dataset contain 6,323, 8,218 and 2,207 shapes, respectively. 

We compare our proposed algorithm with B-Global \cite{zhan2020generative}, B-LSTM \cite{zhan2020generative}, B-Complement \cite{zhan2020generative} and Dynamic Graph Learning (B-DGL) \cite{zhan2020generative}. The results show that our algorithm achieves state-of-the-art performance over other baselines. 

\subsection{Evaluation Metrics}
In our experiments, we follow Huang et al. \cite{zhan2020generative} to use Shape Chamfer Distance (SCD) \cite{zhan2020generative}, Part Accuracy (PA) \cite{li2020learning} and Connectivity Accuracy (CA) \cite{zhan2020generative} to evaluate the assembly quality. SCD and PA are used to evaluate the assembly quality of the whole shape and each individual part, respectively, and CA can show the quality of the connections between each pair of parts. We generate ten shapes for each set of separate parts in the evaluation procedure. Quality evaluations are based on Minimum Matching Distance \cite{dubrovina2019composite}, which means the minimum distance between the assembled shapes and the ground truth is measured for the quality evaluation.

According to Huang et al. \cite{zhan2020generative}, apart from assembly quality, assembly diversity is another crucial aspect of the 3D part assembly task. However, Huang et al. only provide a qualitative evaluation of the diversity of the assembly algorithm in their paper without presenting a quantitative method to assess the diversity. In the following, we introduce our proposed metrics, the Quality-Diversity Score (QDS) and the Weighted Quality-Diversity Score (WQDS).

Diversity Score (DS) \cite{mo2020pt2pc, shu20193d} evaluates the diversity of the results: The formula of DS is 
$\operatorname{DS} = \frac{1}{N^2} \sum_{i,j=1}^N(\operatorname{Dist}(\mathbf{P}^*_i, \mathbf{P}^*_j)),$
where $\mathbf{P}^*_i$ and $\mathbf{P}^*_j$ represent any two assembled shapes. However DS only evaluates the average distances between pairs of transformed shapes but does not consider whether the shapes are reasonably assembled. Therefore, this metric is not suitable for the task of 3D part assembly. To solve this problem, we propose QDS and WQDS that can not only test the diversity among all transformed shapes but also consider the quality of these transformed shapes. The formula is shown in Equation \ref{equ:eval_qds4} and \ref{equ:eval_wqds}. 
We apply SCD as the distance metric $\operatorname{Dist}$ in QDS and WQDS. 
\begin{equation}
\label{equ:eval_qds4}
\operatorname{QDS} = \frac{1}{N^2} \sum_{i,j=1}^N[\operatorname{Dist}(\mathbf{P}^*_i, \mathbf{P}^*_j) \cdot \mathds{1}(\operatorname{CA}(\mathbf{P}^*_i)>\tau_q) \cdot 
\mathds{1} (\operatorname{CA}(\mathbf{P}^*_j)>\tau_q)],
\end{equation}
\begin{equation}
\label{equ:eval_wqds}
\operatorname{WQDS} = \frac{1}{N^2} \sum_{i,j=1}^N[\operatorname{Dist}(\mathbf{P}^*_i, \mathbf{P}^*_j) \cdot \operatorname{CA}(\mathbf{P}^*_i) \cdot 
\operatorname{CA}(\mathbf{P}^*_j)],
\end{equation}
The only difference between QDS/WQDS and DS is that we add constraints to the comparison pair. Specifically, for QDS, the constraint is given by $\mathds{1}(\operatorname{CA}(\mathbf{P}^*_i)>\tau_q) \cdot \mathds{1} (\operatorname{CA}(\mathbf{P}^*_j)>\tau_q)$. In the case of WQDS, the constraint takes the form $\operatorname{CA}(\mathbf{P}^*_i) \cdot 
\operatorname{CA}(\mathbf{P}^*_j)$. As discussed above, CA evaluates the connectivity accuracy of the algorithms. The constraints for the two metrics mean that the pair $\mathbf{P}_i^*$ and $\mathbf{P}_j^*$ contribute to the diversity value if and only if both assembled shapes have sufficiently high connectivity accuracy. In other words, both assembled shapes should have high-quality of connections between each pair of parts. These two new metrics, QDS and WQDS, can meet the requirements of our tasks, which involve evaluating the diversity between reasonably assembled pairs. In our experiments, we set $\tau_q = 0.5$ for QDS.

\vspace{-.3cm}
\subsection{Compare with the Baselines}
\vspace{-.1cm}

\begin{table}[t]
\centering
\small

\setlength{\tabcolsep}{2.45mm}{\begin{tabular}{ccccccc}

\toprule[2pt]
Metrics  & Category  & \multicolumn{1}{c}{B-Global} & \multicolumn{1}{c}{B-LSTM} & \multicolumn{1}{c}{B-Complement} & \multicolumn{1}{c}{B-DGL} & \multicolumn{1}{c}{Ours} \\ \midrule[1pt]
\multirow{3}{*}{SCD ↓}              & Chair     & 0.0178                       & 0.0230                     & 0.0197                           & 0.0089                    & \textbf{0.0071}                  \\ 
                                    & Table     & 0.0077                       & 0.0159                     & 0.0116                           & 0.0051                    & \textbf{0.0042}                  \\  
                                    & Lamp      & 0.0111                       & \textbf{0.0104}            & 0.0157                           & 0.0105                    & 0.0111                           \\ \hline
\multirow{3}{*}{PA ↑}               & Chair     & 13.35                        & 8.92                       & 10.99                            & 38.51                     & \textbf{44.51}                   \\ 
                                    & Table     & 20.01                        & 8.39                       & 15.84                            & 46.57                     & \textbf{52.78}                    \\  
                                    & Lamp      & 13.87                        & 28.24                      & 11.57                            & 33.43                     & \textbf{34.32}                   \\ \hline
\multirow{3}{*}{CA ↑}               & Chair     & 10.01                        & 11.2                       & 10.59                            & 23.51                     & \textbf{30.32}                   \\  
                                    & Table     & 18.08                        & 18.78                      & 15.65                            & 39.63                     & \textbf{40.59}                   \\  
                                    & Lamp      & 27.73                        & 28.67                      & 32.2                             & 40.19                     & \textbf{49.07}                   \\ \hline
      
\multirow{3}{*}{QDS ($10^{-5}$) ↑}  & Chair     & 0.152       & 0.036     & 0.086                   & 1.688    & \textbf{3.355}                  \\ 
                                    & Table     & 0.2        & 0.246      & 0.057           & 3.048    & \textbf{9.172}                  \\  
                                    & Lamp      & 0.758        & 0.629     & 2.814           & 1.835     & \textbf{6.836}                    \\ \hline

\multirow{3}{*}{WQDS ($10^{-4}$) ↑} & Chair     & 0.188       & 0.074     & 0.207                   & 0.553    & \textbf{1.71}                  \\  
                                    & Table     & 0.169       & 0.163     & 0.180           & 0.342    & \textbf{1.8}                  \\  
                                    & Lamp      & 0.175       & 0.211     & 1.0            & 0.31     & \textbf{1.02}                    \\
                                    \bottomrule          
\end{tabular}}
\vspace{.2cm}
\caption{The quantitative comparison between our proposed algorithm and other baselines. In the testing stage, the sequence of the input parts is randomly shuffled. The results show our framework outperforms all the baselines for most metrics (only the SCD score of our framework is slightly below B-LSTM's SCD in Lamp dataset testing). In particular, the QDS of our framework outperforms other baselines by a large margin.}
\vspace{-.5cm}
\label{table:baselines}
\end{table}

Table~\ref{table:baselines} presents the quantitative evaluation results of our algorithm in comparison to other baselines. It is clear to see that our algorithm attains the highest score in most metrics. 
These results indicate that our algorithm can generate diverse and high-quality assembly outcomes. 

Qualitative results are shown in Fig.~\ref{fig:eval_qualitative_results}, which compares both the quality and diversity of our algorithm and other baselines. The results show that only our algorithm can assemble parts in a diverse way while keeping high quality. It is hardly possible for B-Global, B-LSTM, and B-Complement to generate reasonable results. B-DGL can generate some reasonable assembly results, but the generated results lack diversity. The substantial diversity offered by our method implies that Score-PA is also suitable for 3D shape design tasks.

\vspace{-.3cm}
\subsection{Ablation Study}
\label{sec:ablation_study}
\vspace{-.1cm}

\begin{figure}[h]
    \centering
    \vspace{-.2cm}
    \includegraphics[width=1\textwidth]{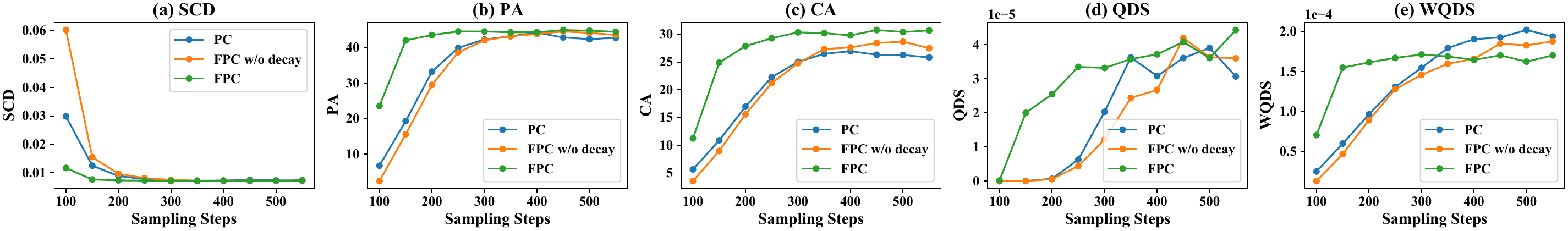}
    \vspace{-.5cm}
  \caption{The quantitative results among PC, FPC without noise decay and FPC. The three samplers are tested with the sampling steps from 100 to 550. The results show that our algorithm has the fastest convergence speed. }
  \label{fig:ablation_fast_sampling}
\end{figure}

Fig.~\ref{fig:ablation_fast_sampling} shows the ablation results among the vanilla PC sampler, the FPC sampler without noise decay and the full algorithm of the FPC sampler. The ablation experiments are conducted on the Chair dataset. We test the three samplers with different sampling steps. 
The results of the five metrics consistently show that the convergence speed of the FPC sampler is much faster than that of the other two algorithms. FPC sampler only requires 200 steps to achieve a relatively high score and 300 steps to achieve optimal performance. 
In comparison, the vanilla PC sampler and the FPC sampler without noise decay need 400 steps and 500 steps, respectively, to achieve similar results to FPC (200 steps). The detailed data can be found in Table \ref{table:optimal_ablation}. Compared with vanilla PC sampler, FPC-200, FPC-250 and 
\begin{wrapfigure}{r}{0.5\textwidth}
    \centering
    \vspace{-.2cm}
    \includegraphics[width=0.5\textwidth,trim=100 70 60 40,clip]{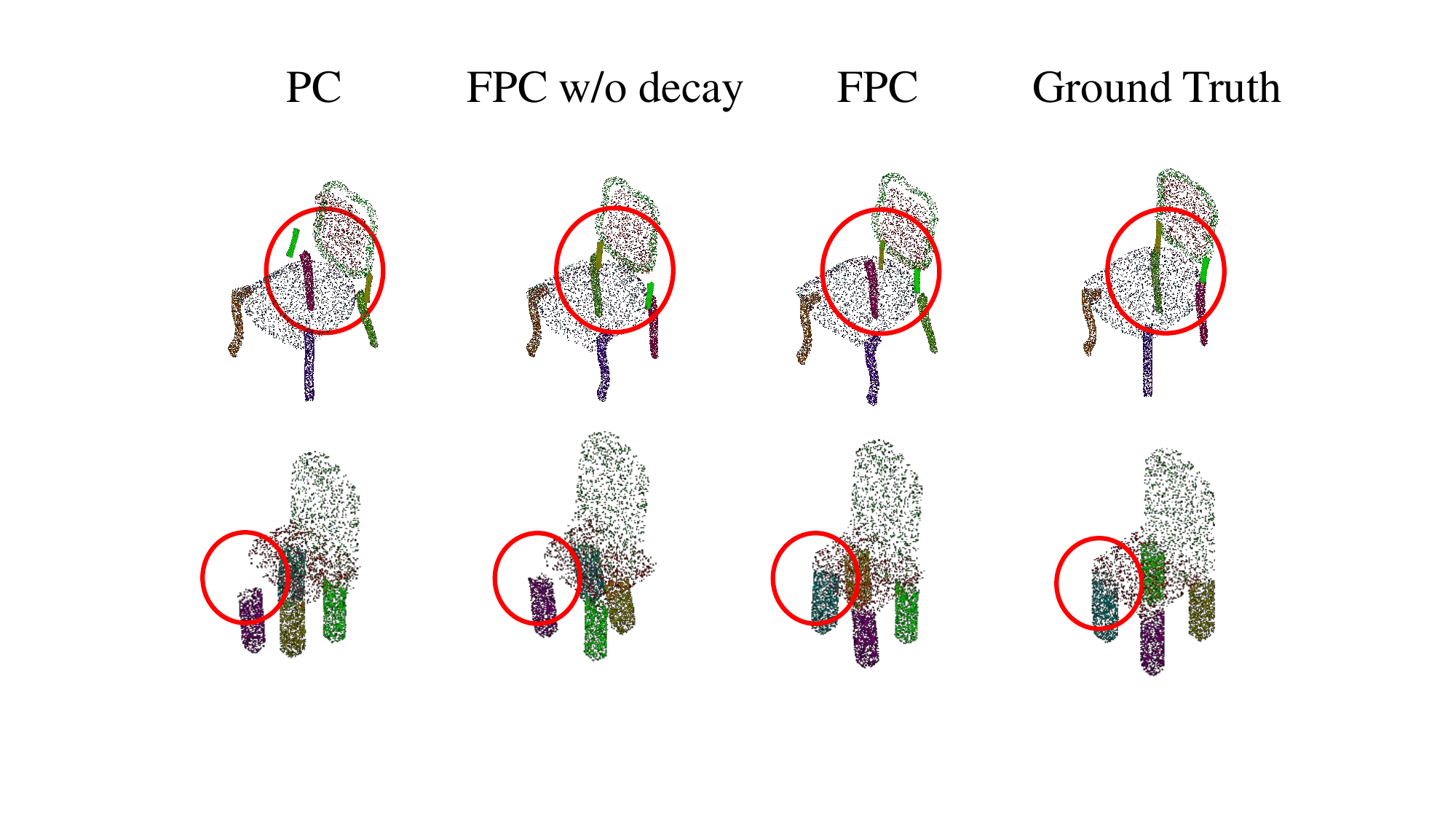}
    \vspace{-.6cm}
  \caption{The qualitative ablation experiments among PC, FPC without noise decay and FPC (all the samplers are tested with their optimal sampling steps). The results sampled by FPC have the best connectivity.}
  \vspace{-.4cm}
  \label{fig:ablation_study}
\end{wrapfigure}
FPC-300 achieve $\times2.27$, $\times1.79$ and $\times1.47$ acceleration respectively. Besides, the comparison between FPC and FPC without noise decay proves that the technique of noise decay can indeed help accelerate the sampling algorithm.

\begin{table}[t]
\centering
\small
\setlength{\tabcolsep}{1.9mm}{\begin{tabular}{ccccccc}
\toprule[1pt]
Samplers  & SCD ↓             & PA ↑              & CA ↑           & QDS ($10^{-5}$) ↑   &    WQDS ($10^{-4}$) ↑       & Avg. Time (s) ↓        \\ \midrule[.7pt]
PC            & 0.0073          & 44.22          & 26.97          & 3.078          & \textbf{1.902}  & 0.84 ($\times1$)    \\ 
FPC w/o decay & 0.0072           & 44.13          & 28.65          & \textbf{3.629}   & 1.825      & 0.99 ($\times0.85$) \\ 
FPC-200       & 0.0073          & 43.52          & 27.88          & 2.545          & 1.612   & \textbf{0.37 ($\times2.27$)} \\
FPC-250       & 0.0073          & \textbf{44.53} & 29.27          & 3.345          & 1.668     & 0.47 ($\times1.79$)   \\ 
FPC-300       & \textbf{0.0071} & 44.51          & \textbf{30.32} & 3.355          & 1.71   & 0.57 ($\times1.47$)   \\ \bottomrule
\end{tabular}}
\vspace{.2cm}
\caption{ Compare FPC (200, 250 and 300 sampling steps) with the optimal PC and FPC without noise decay sampler. All the samplers are tested in the same hardware environment (R9 3900x with RTX 3090).}
\vspace{-.4cm}
\label{table:optimal_ablation}
\end{table}

\textbf{Connectivity accuracy enhancement.} 
Our proposed FPC algorithm also has a good performance on connectivity accuracy, which can be proved by both the results shown in Fig.~\ref{fig:ablation_fast_sampling} and the qualitative results shown in Fig.~\ref{fig:ablation_study}. Fig.~\ref{fig:ablation_study} shows that the results sampled by PC or FPC without noise decay are easier to be disconnected (see the red circles in Fig.~\ref{fig:ablation_study}), while this does not appear in the results sampled by FPC sampler. 

\section{Conclusion}
\label{sec:conclusion}
In this work, we propose a novel framework, Score-PA, for the 3D part assembly task, viewing the 3D part assembly as a problem of conditional probability distribution estimation. We modify the original score-matching objective function \cite{song2020score} for log-conditional-density estimation purposes, and develop a graph neural network for score function modelling. Besides, we propose a new sampling method, FPC, to speed up the inference of our framework. The experiments demonstrate that our designed framework achieves the current state-of-the-art performance over other baselines for both assembly quality and assembly diversity.

\vspace{-0.5cm}
\section*{Acknowledgement}
\vspace{-0.2cm}
This project was supported by the National Natural Science Foundation of China - General Program (62376006) and The National Youth Talent Support Program (8200800081).
\vspace{-0.2cm}
\section{Appendix}

\appendix

\section{More Discussion about Our Score-PA}

\paragraph{Selection of the weight function $\lambda(t)$} We discuss our training algorithm in the main text. $\lambda(t)$ is important for our training objective function. According to Song et al. \cite{song2020score}, we need to choose a suitable $\lambda(t)$ to make our prior distribution $p(\mathbf{Q_P}(T))$ independent from the data distribution and easily sampled (We set $T=1$). In our algorithm, the weight function is selected as $\lambda(t) = \frac{1}{2 \log \sigma}(\sigma^{2t} - 1)$, which follows Song et al.'s setting \cite{song2020score}. We already have our SDE $d \mathbf{Q_P} = \sigma^t d\mathbf{w}$, where $t\in[0,1]$, and in this situation, 
\begin{equation}
p_{0t}(\mathbf{Q_P}(t) \mid \mathbf{Q_P}(0), \mathbf{P}) = \mathcal{N}\bigg(\mathbf{Q_P}(t); \mathbf{Q_P}(0), \frac{1}{2\log \sigma}(\sigma^{2t} - 1) \mathbf{I}\bigg)
\end{equation}
We have our weight function $\lambda(t) = \frac{1}{2 \log \sigma}(\sigma^{2t} - 1)$, and then the prior distribution $p_{t=1}$ is,
\begin{equation}
\int p_0(\mathbf{y})\mathcal{N}\bigg(\mathbf{Q_P}; \mathbf{y}, \frac{1}{2 \log \sigma}(\sigma^2 - 1)\mathbf{I}\bigg) d \mathbf{y} \approx \mathcal{N}\bigg(\mathbf{Q_P}; \mathbf{0}, \frac{1}{2 \log \sigma}(\sigma^2 - 1)\mathbf{I}\bigg),
\label{Equ:prior}
\end{equation}
where $\sigma$ should be large enough. The Equation \ref{Equ:prior} means that the prior distribution can be easily sampled from a normal distribution. 

\section{More Qualitative Comparisons}
\begin{figure}[t]
  \centering
  \includegraphics[width=1.0\textwidth,trim=40 200 150 10,clip]{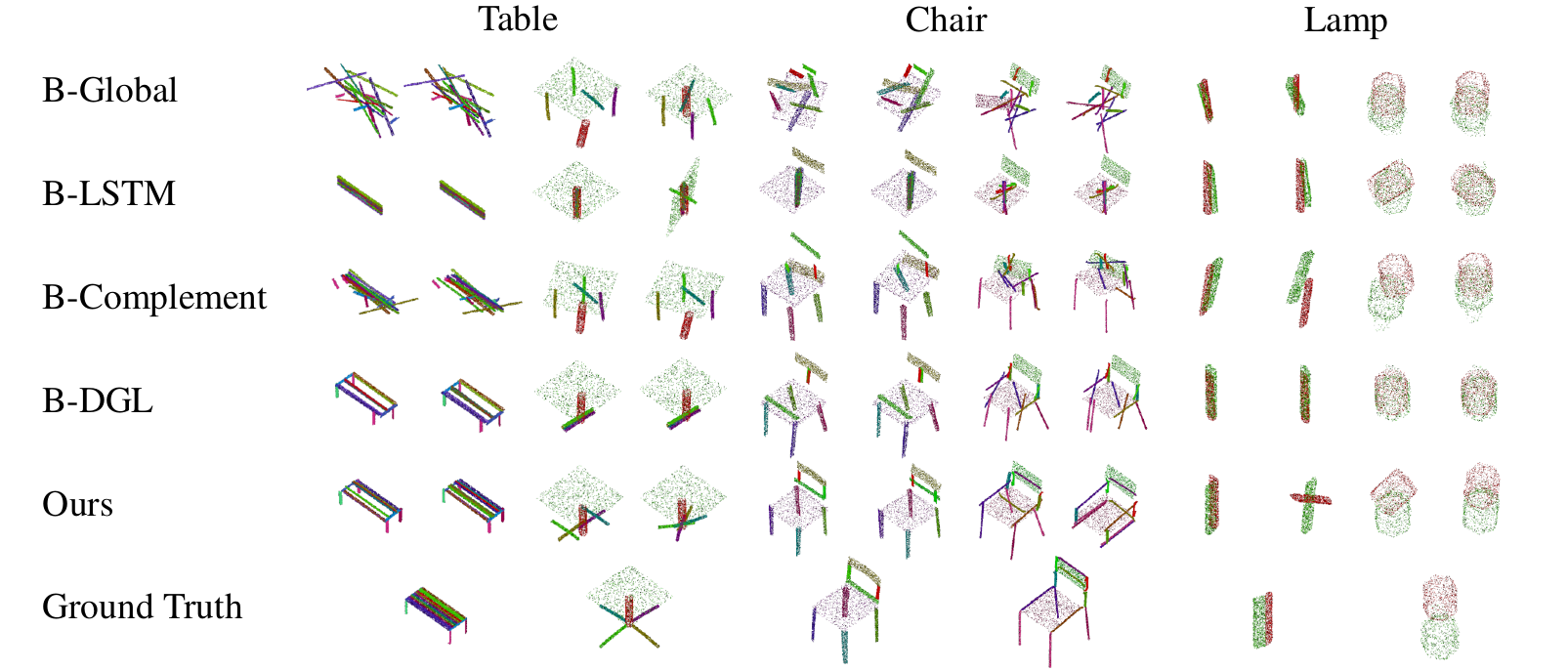}
  \caption{More qualitative comparisons between our algorithm and other baselines. }
  \vspace{-.4cm}
  \label{fig:supp_results}
\end{figure}
We present more qualitative comparisons between our algorithm and other baselines in Figure \ref{fig:supp_results}. Similar to the comparisons in the main text, we present two assembly results per input parts $\mathbf{P}$ for each algorithm. The comparisons show that only our algorithm is able to generate diverse results with high quality. 

\section{Details about Our Experiments}
\paragraph{Training details} As discussed in Section 3 of our paper, our algorithm has two important hyper-parameters $T$ and $\sigma$ in the training procedure. In our experiments on the three datasets, we set $T=1.0$, and $\sigma=25.0$. We train our models with 2000 Epochs on Chair and Table datasets, and 4000 Epochs on Lamp dataset. The learning rate for all datasets are set as $10^{-4}$, and the Optimizer is Adam. The training batch size for all the datasets is 16. 

\paragraph{Other details} We conducted both training and testing experiments using a single RTX 3090 GPU. To ensure reproducibility, we set a fixed random seed. In our ablation study, we define the sampling steps for FPC and FPC w/o decay as $steps = N + C_F$. For the PC sampler, the number of sampling steps is simply $N$, as it does not involve a decay stage. In all testing experiments, including the ablation study, we set the sampling batch size for Score-PA to 4.

\section{Limitation and Future Work}
Currently, we achieve diverse part assembly in an ideal simulation environment. In the future, we plan to take the physical factors (\emph{e.g.}, physical collision) into our consideration, and achieve autonomous part assembly in a real physical environment. 

\bibliography{egbib}

\begin{thebibliography}{31}
\providecommand{\natexlab}[1]{#1}
\providecommand{\url}[1]{\texttt{#1}}
\expandafter\ifx\csname urlstyle\endcsname\relax
  \providecommand{\doi}[1]{doi: #1}\else
  \providecommand{\doi}{doi: \begingroup \urlstyle{rm}\Url}\fi

\bibitem[Brooks(1998)]{brooks1998markov}
Stephen Brooks.
\newblock Markov chain monte carlo method and its application.
\newblock \emph{Journal of the royal statistical society: series D (the Statistician)}, 47\penalty0 (1):\penalty0 69--100, 1998.

\bibitem[Brooks et~al.(2011)Brooks, Gelman, Jones, and Meng]{brooks2011handbook}
Steve Brooks, Andrew Gelman, Galin Jones, and Xiao-Li Meng.
\newblock \emph{Handbook of markov chain monte carlo}.
\newblock CRC press, 2011.

\bibitem[Cai et~al.(2020)Cai, Yang, Averbuch-Elor, Hao, Belongie, Snavely, and Hariharan]{cai2020learning}
Ruojin Cai, Guandao Yang, Hadar Averbuch-Elor, Zekun Hao, Serge Belongie, Noah Snavely, and Bharath Hariharan.
\newblock Learning gradient fields for shape generation.
\newblock In \emph{European Conference on Computer Vision}, pages 364--381. Springer, 2020.

\bibitem[Carlo(2004)]{carlo2004markov}
Chain~Monte Carlo.
\newblock Markov chain monte carlo and gibbs sampling.
\newblock \emph{Lecture notes for EEB}, 581:\penalty0 540, 2004.

\bibitem[Chaudhuri et~al.(2011)Chaudhuri, Kalogerakis, Guibas, and Koltun]{chaudhuri2011probabilistic}
Siddhartha Chaudhuri, Evangelos Kalogerakis, Leonidas Guibas, and Vladlen Koltun.
\newblock Probabilistic reasoning for assembly-based 3d modeling.
\newblock In \emph{ACM SIGGRAPH 2011 papers}, pages 1--10. 2011.

\bibitem[Ci et~al.(2022)Ci, Wu, Zhu, Ma, Dong, Zhong, and Wang]{ci2022gfpose}
Hai Ci, Mingdong Wu, Wentao Zhu, Xiaoxuan Ma, Hao Dong, Fangwei Zhong, and Yizhou Wang.
\newblock Gfpose: Learning 3d human pose prior with gradient fields.
\newblock \emph{arXiv preprint arXiv:2212.08641}, 2022.

\bibitem[Dubrovina et~al.(2019)Dubrovina, Xia, Achlioptas, Shalah, Groscot, and Guibas]{dubrovina2019composite}
Anastasia Dubrovina, Fei Xia, Panos Achlioptas, Mira Shalah, Rapha{\"e}l Groscot, and Leonidas~J Guibas.
\newblock Composite shape modeling via latent space factorization.
\newblock In \emph{Proceedings of the IEEE/CVF International Conference on Computer Vision}, pages 8140--8149, 2019.

\bibitem[Funkhouser et~al.(2004)Funkhouser, Kazhdan, Shilane, Min, Kiefer, Tal, Rusinkiewicz, and Dobkin]{funkhouser2004modeling}
Thomas Funkhouser, Michael Kazhdan, Philip Shilane, Patrick Min, William Kiefer, Ayellet Tal, Szymon Rusinkiewicz, and David Dobkin.
\newblock Modeling by example.
\newblock \emph{ACM transactions on graphics (TOG)}, 23\penalty0 (3):\penalty0 652--663, 2004.

\bibitem[Geyer(1992)]{geyer1992practical}
Charles~J Geyer.
\newblock Practical markov chain monte carlo.
\newblock \emph{Statistical science}, pages 473--483, 1992.

\bibitem[Harish et~al.(2021)Harish, Nagar, and Raman]{harish2021rgl}
Abhinav~Narayan Harish, Rajendra Nagar, and Shanmuganathan Raman.
\newblock Rgl-net: A recurrent graph learning framework for progressive part assembly.
\newblock \emph{arXiv preprint arXiv:2107.12859}, 2021.

\bibitem[Huang et~al.(2020)Huang, Zhan, Fan, Mo, Shao, Chen, Guibas, and Dong]{zhan2020generative}
Jialei Huang, Guanqi Zhan, Qingnan Fan, Kaichun Mo, Lin Shao, Baoquan Chen, Leonidas~J Guibas, and Hao Dong.
\newblock Generative 3d part assembly via dynamic graph learning.
\newblock \emph{Advances in Neural Information Processing Systems}, 33:\penalty0 6315--6326, 2020.

\bibitem[Hyv{\"a}rinen and Dayan(2005)]{hyvarinen2005estimation}
Aapo Hyv{\"a}rinen and Peter Dayan.
\newblock Estimation of non-normalized statistical models by score matching.
\newblock \emph{Journal of Machine Learning Research}, 6\penalty0 (4), 2005.

\bibitem[Jaiswal et~al.(2016)Jaiswal, Huang, and Rai]{jaiswal2016assembly}
Prakhar Jaiswal, Jinmiao Huang, and Rahul Rai.
\newblock Assembly-based conceptual 3d modeling with unlabeled components using probabilistic factor graph.
\newblock \emph{Computer-Aided Design}, 74:\penalty0 45--54, 2016.

\bibitem[Kalogerakis et~al.(2012)Kalogerakis, Chaudhuri, Koller, and Koltun]{kalogerakis2012probabilistic}
Evangelos Kalogerakis, Siddhartha Chaudhuri, Daphne Koller, and Vladlen Koltun.
\newblock A probabilistic model for component-based shape synthesis.
\newblock \emph{Acm Transactions on Graphics (TOG)}, 31\penalty0 (4):\penalty0 1--11, 2012.

\bibitem[Li et~al.(2020)Li, Mo, Shao, Sung, and Guibas]{li2020learning}
Yichen Li, Kaichun Mo, Lin Shao, Minhyuk Sung, and Leonidas Guibas.
\newblock Learning 3d part assembly from a single image.
\newblock In \emph{European Conference on Computer Vision}, pages 664--682. Springer, 2020.

\bibitem[Luo and Hu(2021)]{luo2021score}
Shitong Luo and Wei Hu.
\newblock Score-based point cloud denoising.
\newblock In \emph{Proceedings of the IEEE/CVF International Conference on Computer Vision}, pages 4583--4592, 2021.

\bibitem[Mo et~al.(2019)Mo, Zhu, Chang, Yi, Tripathi, Guibas, and Su]{mo2019partnet}
Kaichun Mo, Shilin Zhu, Angel~X Chang, Li~Yi, Subarna Tripathi, Leonidas~J Guibas, and Hao Su.
\newblock Partnet: A large-scale benchmark for fine-grained and hierarchical part-level 3d object understanding.
\newblock In \emph{Proceedings of the IEEE/CVF conference on computer vision and pattern recognition}, pages 909--918, 2019.

\bibitem[Mo et~al.(2020)Mo, Wang, Yan, and Guibas]{mo2020pt2pc}
Kaichun Mo, He~Wang, Xinchen Yan, and Leonidas Guibas.
\newblock Pt2pc: Learning to generate 3d point cloud shapes from part tree conditions.
\newblock In \emph{European Conference on Computer Vision}, pages 683--701. Springer, 2020.

\bibitem[Qi et~al.(2017)Qi, Su, Mo, and Guibas]{qi2017pointnet}
Charles~R Qi, Hao Su, Kaichun Mo, and Leonidas~J Guibas.
\newblock Pointnet: Deep learning on point sets for 3d classification and segmentation.
\newblock In \emph{Proceedings of the IEEE conference on computer vision and pattern recognition}, pages 652--660, 2017.

\bibitem[Schor et~al.(2019)Schor, Katzir, Zhang, and Cohen-Or]{schor2019componet}
Nadav Schor, Oren Katzir, Hao Zhang, and Daniel Cohen-Or.
\newblock Componet: Learning to generate the unseen by part synthesis and composition.
\newblock In \emph{Proceedings of the IEEE/CVF International Conference on Computer Vision}, pages 8759--8768, 2019.

\bibitem[Shi et~al.(2021)Shi, Luo, Xu, and Tang]{shi2021learning}
Chence Shi, Shitong Luo, Minkai Xu, and Jian Tang.
\newblock Learning gradient fields for molecular conformation generation.
\newblock In \emph{International Conference on Machine Learning}, pages 9558--9568. PMLR, 2021.

\bibitem[Shu et~al.(2019)Shu, Park, and Kwon]{shu20193d}
Dong~Wook Shu, Sung~Woo Park, and Junseok Kwon.
\newblock 3d point cloud generative adversarial network based on tree structured graph convolutions.
\newblock In \emph{Proceedings of the IEEE/CVF International Conference on Computer Vision}, pages 3859--3868, 2019.

\bibitem[Song and Ermon(2019)]{song2019generative}
Yang Song and Stefano Ermon.
\newblock Generative modeling by estimating gradients of the data distribution.
\newblock \emph{Advances in Neural Information Processing Systems}, 32, 2019.

\bibitem[Song and Ermon(2020)]{song2020improved}
Yang Song and Stefano Ermon.
\newblock Improved techniques for training score-based generative models.
\newblock \emph{Advances in neural information processing systems}, 33:\penalty0 12438--12448, 2020.

\bibitem[Song et~al.(2020{\natexlab{a}})Song, Garg, Shi, and Ermon]{song2020sliced}
Yang Song, Sahaj Garg, Jiaxin Shi, and Stefano Ermon.
\newblock Sliced score matching: A scalable approach to density and score estimation.
\newblock In \emph{Uncertainty in Artificial Intelligence}, pages 574--584. PMLR, 2020{\natexlab{a}}.

\bibitem[Song et~al.(2020{\natexlab{b}})Song, Sohl-Dickstein, Kingma, Kumar, Ermon, and Poole]{song2020score}
Yang Song, Jascha Sohl-Dickstein, Diederik~P Kingma, Abhishek Kumar, Stefano Ermon, and Ben Poole.
\newblock Score-based generative modeling through stochastic differential equations.
\newblock \emph{arXiv preprint arXiv:2011.13456}, 2020{\natexlab{b}}.

\bibitem[Song et~al.(2021)Song, Durkan, Murray, and Ermon]{song2021maximum}
Yang Song, Conor Durkan, Iain Murray, and Stefano Ermon.
\newblock Maximum likelihood training of score-based diffusion models.
\newblock \emph{Advances in Neural Information Processing Systems}, 34:\penalty0 1415--1428, 2021.

\bibitem[Suhail et~al.(2021)Suhail, Mittal, Siddiquie, Broaddus, Eledath, Medioni, and Sigal]{suhail2021energy}
Mohammed Suhail, Abhay Mittal, Behjat Siddiquie, Chris Broaddus, Jayan Eledath, Gerard Medioni, and Leonid Sigal.
\newblock Energy-based learning for scene graph generation.
\newblock In \emph{Proceedings of the IEEE/CVF Conference on Computer Vision and Pattern Recognition}, pages 13936--13945, 2021.

\bibitem[Tancik et~al.(2020)Tancik, Srinivasan, Mildenhall, Fridovich-Keil, Raghavan, Singhal, Ramamoorthi, Barron, and Ng]{tancik2020fourier}
Matthew Tancik, Pratul Srinivasan, Ben Mildenhall, Sara Fridovich-Keil, Nithin Raghavan, Utkarsh Singhal, Ravi Ramamoorthi, Jonathan Barron, and Ren Ng.
\newblock Fourier features let networks learn high frequency functions in low dimensional domains.
\newblock \emph{Advances in Neural Information Processing Systems}, 33:\penalty0 7537--7547, 2020.

\bibitem[Wu et~al.(2022)Wu, Zhong, Xia, and Dong]{wu2022targf}
Mingdong Wu, Fangwei Zhong, Yulong Xia, and Hao Dong.
\newblock Tar{GF}: Learning target gradient field for object rearrangement.
\newblock In Alice~H. Oh, Alekh Agarwal, Danielle Belgrave, and Kyunghyun Cho, editors, \emph{Advances in Neural Information Processing Systems}, 2022.
\newblock URL \url{https://openreview.net/forum?id=Euv1nXN98P3}.

\bibitem[Wu et~al.(2020)Wu, Zhuang, Xu, Zhang, and Chen]{wu2020pq}
Rundi Wu, Yixin Zhuang, Kai Xu, Hao Zhang, and Baoquan Chen.
\newblock Pq-net: A generative part seq2seq network for 3d shapes.
\newblock In \emph{Proceedings of the IEEE/CVF Conference on Computer Vision and Pattern Recognition}, pages 829--838, 2020.

\end{thebibliography}
\end{document}